\documentclass{article}
\usepackage{ijcai17}

\usepackage{times}
\usepackage{helvet}
\usepackage{courier}
\usepackage{multirow}
\usepackage{graphicx}
\usepackage{amsfonts}
\usepackage{stmaryrd}
\usepackage{bm}
\usepackage{amsfonts}
\usepackage{algorithm}
\usepackage{algpseudocode}
\usepackage{color}

\def\argmax{\mathop{\rm argmax}}%
\title{Joint Training for Pivot-based Neural Machine Translation}
\author{Yong Cheng$^\#$, Yang Liu$^\dagger$, Qian Yang$^\#$,  Maosong Sun$^\dagger$ and  Wei Xu$^\#$ \\
 $^\#$Institute for Interdisciplinary Information Sciences, Tsinghua University, Beijing, China  \\
 $^\dagger$State Key Laboratory of Intelligent Technology and Systems  \\
  Tsinghua National Laboratory for Information Science and Technology \\
 Department of Computer Science and Technology, Tsinghua University, Beijing, China\\
 {\tt chengyong3001@gmail.com,qian-yang12@mails.tsinghua.edu.cn } \\
 {\tt \{liuyang2011, weixu, sms\}@tsinghua.edu.cn } \\
}

\begin{document}

\maketitle

\begin{abstract}
While recent neural machine translation approaches have delivered state-of-the-art performance for resource-rich language pairs, they suffer from the data scarcity problem for resource-scarce language pairs. Although this problem can be alleviated by exploiting a pivot language to bridge the source and target languages, the source-to-pivot and pivot-to-target translation models are usually independently trained. In this work, we introduce a joint training algorithm for pivot-based neural machine translation. We propose three methods to connect the two models and enable them to interact with each other during training. Experiments on Europarl and WMT corpora show that joint training of source-to-pivot and pivot-to-target models leads to significant improvements over independent training across various languages.

\end{abstract}

\section{Introduction}
Recent several years have witnessed the rapid development of neural machine translation (NMT) \cite{Sutskever:14,Bahdanau:15}, which advocates the use of neural networks to directly model the translation process in an end-to-end way. Thanks to the capability of learning representations from training data, NMT systems have achieved significant improvements over conventional statistical machine translation (SMT) across a variety of language pairs \cite{Junczys-Dowmunt:16,johnson:16}.

However, there still remains a major challenge for NMT: large-scale parallel corpora are usually non-existent for most language pairs. This is unfortunate because NMT is a data-hungry approach and requires a large amount of data to fully train model parameters. Without sufficient training data, NMT tends to learn poor estimates on low-count events. Zoph {\em et al}. \shortcite{zoph:16} indicate that NMT obtains much worse translation quality than SMT when only small-scale parallel corpora are available.

As a result, improving neural machine translation on resource-scarce language pairs has attracted much attention in the community \cite{firat:16,zoph:16,johnson:16}. Most existing methods focus on leveraging data of multiple resource-rich language pairs to improve NMT for resource-scarce language pairs. Firat {\em et al}. \shortcite{firat:16} propose multi-way, multilingual neural machine translation to achieve direct source-to-target translation even without parallel data available. Zoph {\em et al}. \shortcite{zoph:16} present a transfer learning method that transfers the model parameters trained for resource-rich language pairs to initialize and constrain the translation model training of resource-scarce language pairs. Johnson {\em et al.} \shortcite{johnson:16} introduce a universal NMT model for all language pairs, which takes advantage of multilingual data to improve NMT for all languages involved.

Bridging source and target languages with a {\em pivot} language is another important direction, which has been intensively studied in conventional SMT \cite{cohn:07,wu:07,utiyama:07,bertoldi:08,zahabi:13,el:13}. Pivot-based approaches assume that there exist source-pivot and pivot-target parallel corpora, which can be used to train source-to-pivot and pivot-to-target translation models, respectively.  One of the most representative approaches, triangulation approach, is to construct a source-to-target phrase table through combining source-to-pivot and pivot-to-target phrase tables.
Another representative approach adopts a pivot-based translation strategy. As a result, source-to-target translation can be divided into two steps: the source sentence is first translated into a pivot sentence using the source-to-pivot model, which is then translated to a target sentence using the pivot-to-target model. Pivot-based approaches have been widely used in SMT due to its simplicity, effectiveness, and minimum requirement of multilingual data. Recently, Johnson {\em et al}. \shortcite{johnson:16} adapt pivot-based approaches to NMT and show that their universal model without incremental training achieves much worse translation performance than pivot-based NMT.

However, pivot-based approaches often suffer from the error propagation problem: the errors made in the source-to-pivot translation will be propagated to the pivot-to-target translation. This can be partly attributed to the discrepancy between source-pivot and pivot-target parallel corpora since they are usually loosely-related or even unrelated. To aggregate the situation, source-to-pivot and pivot-to-target translation models are trained independently, which further enlarges the gap between source and target languages.

In this work, we propose an approach to joint training for pivot-based neural machine translation. The basic idea is to connect the source-to-pivot and pivot-to-target NMT models and enable them to interact with each other during training. This can be done either by encouraging the sharing of word embeddings on the pivot language or by maximizing the likelihood of the cascaded model on a small source-target parallel corpus. Experiments on the Europarl and WMT corpora show that joint training of source-to-pivot and pivot-to-target models obtains significant improvements over independent training.

\section{Background}

Given a source language sentence $\mathbf{x}$ and a target language sentence $\mathbf{y}$, we use $P(\mathbf{y}|\mathbf{x}; \bm{\theta}_{x \rightarrow y})$ to denote a standard attention-based neural machine translation model \cite{Bahdanau:15}, where $\bm{\theta}_{x \rightarrow y}$ is a set of model parameters.

Ideally, the source-to-target model can be trained on a source-target parallel corpus $D_{x,y}=\{\langle \mathbf{x}^{(s)}, \mathbf{y}^{(s)} \rangle\}_{s=1}^{S}$ using maximum likelihood estimation:
\begin{eqnarray}
\hat{\bm{\theta}}_{x \rightarrow y} = \argmax_{\bm{\theta}_{x \rightarrow y}}\Big\{ \mathcal{L}(\bm{\theta}_{x \rightarrow y}) \Big\}
\end{eqnarray}
where the log-likelihood is defined as
\begin{eqnarray}
\mathcal{L}(\bm{\theta}_{x \rightarrow y}) = \sum_{s=1}^{S} \log P(\mathbf{y}^{(s)}|\mathbf{x}^{(s)}; \bm{\theta}_{x \rightarrow y})
\end{eqnarray}

Unfortunately, parallel corpora are usually not readily available for low-resource language pairs. Instead, one can assume that there exist a third language called {\em pivot} with source-pivot and pivot-target parallel corpora available. As a result, it is possible to bridge the source and target languages with the pivot \cite{cohn:07,wu:07,utiyama:07,bertoldi:08,zahabi:13,el:13}.

Let $\mathbf{z}$ be a pivot language sentence. The source-to-target model can be decomposed into two sub-models by treating the pivot sentence as a latent variable:
\begin{eqnarray}
&& P(\mathbf{y}|\mathbf{x}; \bm{\theta}_{x \rightarrow z}, \bm{\theta}_{z \rightarrow y}) \nonumber \\
&=& \sum_{\mathbf{z}} P(\mathbf{z}|\mathbf{x}; \bm{\theta}_{x \rightarrow z}) P(\mathbf{y}|\mathbf{z}; \bm{\theta}_{z \rightarrow y}) \label{eq:source-to-target}
\end{eqnarray}

Let $D_{x,z}=\{\langle \mathbf{x}^{(m)}, \mathbf{z}^{(m)} \rangle\}_{m=1}^{M}$ be a source-pivot parallel corpus, and $D_{z,y}=\{\langle \mathbf{z}^{(n)}, \mathbf{y}^{(n)} \rangle\}_{n=1}^{N}$ be a pivot-target parallel corpus. The source-to-pivot and pivot-to-target models can be {\bf independently} trained on the two parallel corpora, respectively:
\begin{eqnarray}
\hat{\bm{\theta}}_{x \rightarrow z} = \argmax_{\bm{\theta}_{x \rightarrow z}}\Big\{ \mathcal{L}(\bm{\theta}_{x \rightarrow z}) \Big\} \\
\hat{\bm{\theta}}_{z \rightarrow y} = \argmax_{\bm{\theta}_{z \rightarrow y}}\Big\{ \mathcal{L}(\bm{\theta}_{z \rightarrow y}) \Big\}
\end{eqnarray}
where the log-likelihoods are defined as:
\begin{eqnarray}
\mathcal{L}(\bm{\theta}_{x \rightarrow z}) &=& \sum_{m=1}^{M}\log P(\mathbf{z}^{(m)}|\mathbf{x}^{(m)}; \bm{\theta}_{x \rightarrow z}) \\
\mathcal{L}(\bm{\theta}_{z \rightarrow y}) &=& \sum_{n=1}^{N}\log P(\mathbf{y}^{(n)}|\mathbf{z}^{(n)}; \bm{\theta}_{z \rightarrow y})
\end{eqnarray}

\begin{figure}[!t]
\centering
\includegraphics[width=0.48\textwidth]{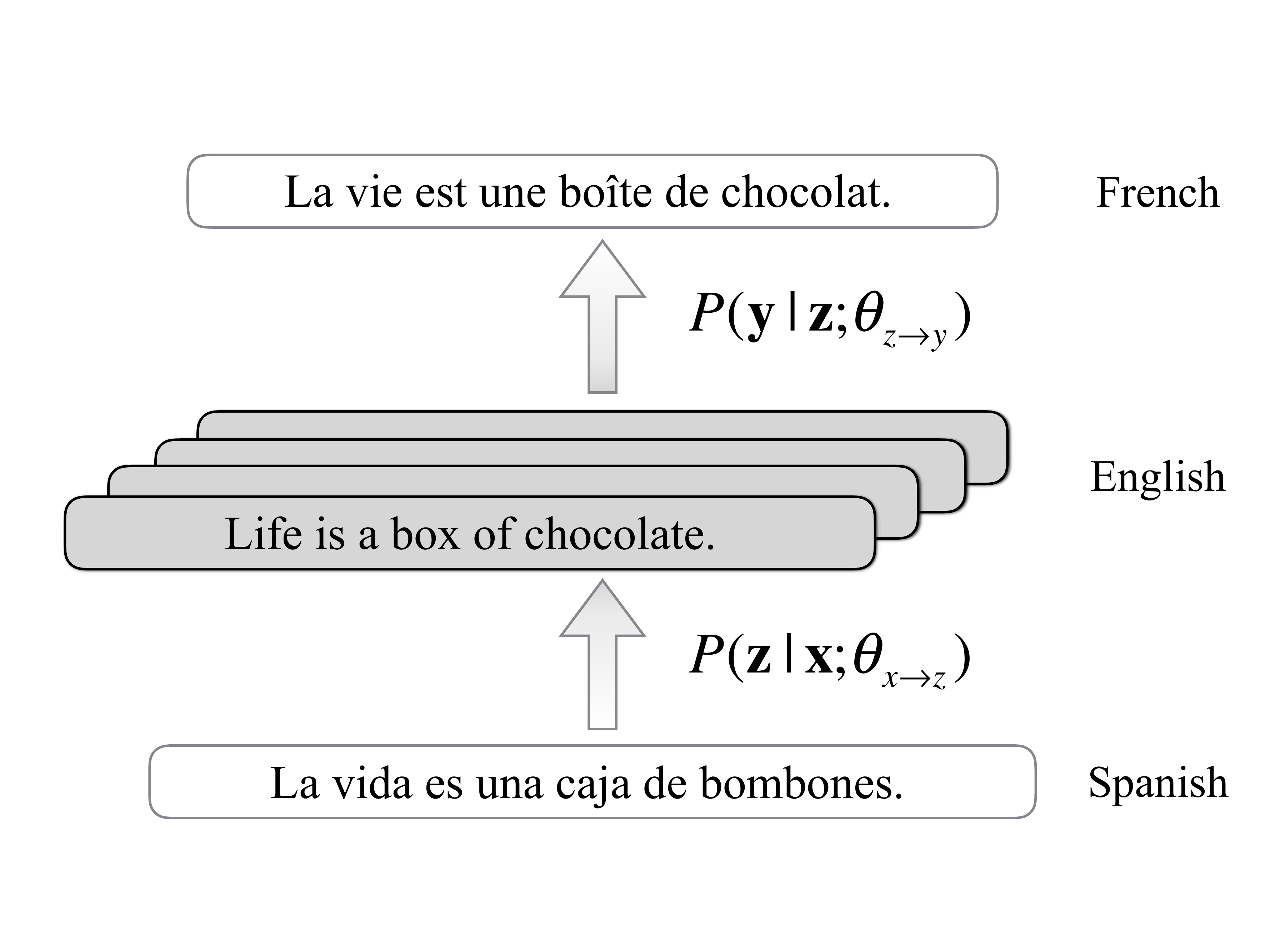}
\caption{The illustration of translation on Spanish-French with English as the pivot language. The Spanish-English NMT model $P(\mathbf{z}|\mathbf{x};\bm{\theta}_{x \rightarrow z})$ first transforms a Spanish sentence into latent English sentences, from which English-French NMT model $P(\mathbf{y}|\mathbf{z};\bm{\theta}_{z \rightarrow y})$ attempts to generate a French sentence corresponding to the Spanish sentence.  } \label{fig:example}
\end{figure}

As Figure \ref{fig:example} shows, a pivot-based translation strategy is usually adopted. Given an unseen source sentence to be translated $\mathbf{x}$, the decision rule is given by:
\begin{eqnarray}
\hat{\mathbf{y}} = \argmax_{\mathbf{y}}\Bigg\{ \sum_{\mathbf{z}}P(\mathbf{z}|\mathbf{x}; \hat{\bm{\theta}}_{x \rightarrow z}) P(\mathbf{y}|\mathbf{z}; \hat{\bm{\theta}}_{z \rightarrow y}) \Bigg\}
\end{eqnarray}

Due to the exponential search space of the pivot language, the decoding process is usually approximated with two steps. The first step translates the source sentence $\mathbf{x}$ into a pivot sentence:
\begin{eqnarray}
\hat{\mathbf{z}} = \argmax_{\mathbf{z}}\Big\{ P(\mathbf{z}|\mathbf{x}; \hat{\bm{\theta}}_{x \rightarrow z}) \Big\} \label{eq:s2p_decoding}
\end{eqnarray}

Then, the pivot sentence is translated to a target sentence:
\begin{eqnarray}
\hat{\mathbf{y}} = \argmax_{\mathbf{y}}\Big\{ P(\mathbf{y}|\hat{\mathbf{z}}; \hat{\bm{\theta}}_{z \rightarrow y}) \Big\} \label{eq:p2t_decoding}
\end{eqnarray}

Although pivot-based approaches are widely for addressing the data scarcity problem in machine translation, they suffer from cascaded translation errors: the mistakes made in the source-to-pivot translation as shown in Eq. (\ref{eq:s2p_decoding}) will be propagated to the pivot-to-target translation as shown in Eq. (\ref{eq:p2t_decoding}). This can be partly attributed to the {\bf model discrepancy} problem: the source-to-pivot and pivot-to-target models are quite different in terms of vocabulary and parameter space because the source-pivot and pivot-target parallel corpora are usually loosely-related or even unrelated. To make things worse, the source-to-pivot model $P(\mathbf{z}|\mathbf{x}; \bm{\theta}_{x \rightarrow z})$ and the pivot-to-target model $P(\mathbf{y}|\mathbf{z}; \bm{\theta}_{z \rightarrow y})$ are trained on the two parallel corpora {\bf independently}, which further increases the discrepancy between two models.

Therefore, it is important to reduce the discrepancy between source-to-pivot and pivot-to-target models to further improve pivot-based neural machine translation.

\section{Joint Training for Pivot-based NMT}

\subsection{Training Objective}

To alleviate the model discrepancy problem, we propose an approach to joint training for pivot-based neural machine translation. The basic idea is to connect source-to-pivot and pivot-to-target models and enable them to interact with each other during training. Our new training objective is given by:
\begin{eqnarray}
&&\mathcal{J}(\bm{\theta}_{x \rightarrow z}, \bm{\theta}_{z \rightarrow y}) \nonumber \\
&=& \mathcal{L}(\bm{\theta}_{x \rightarrow z}) + \mathcal{L}(\bm{\theta}_{z \rightarrow y})+ \lambda \mathcal{R}(\bm{\theta}_{x \rightarrow z}, \bm{\theta}_{z \rightarrow y})
\end{eqnarray}
Note that the training objective consists of three parts: the source-to-pivot likelihood $\mathcal{L}(\bm{\theta}_{x \rightarrow z})$, the pivot-to-target likelihood $\mathcal{L}(\bm{\theta}_{z \rightarrow y})$, and a connection term $\mathcal{R}(\bm{\theta}_{x \rightarrow z}, \bm{\theta}_{z \rightarrow y})$. The hyper-parameter $\lambda$ is used to balance the preference between likelihoods and the connection term.

We expect that the connection term associates the source-to-pivot model $\bm{\theta}_{x \rightarrow z}$ with the pivot-to-target model $\bm{\theta}_{z \rightarrow y}$ and enables the interaction between two models during
training. In the following subsection, we will introduce the three connection terms used in our experiments.

\subsection{Connection Terms}

It is difficult to connect the source-to-pivot and pivot-to-target models during training because the source-to-pivot and pivot-to-target models are distantly-related by definition. More importantly, NMT lacks linguistically interpretable language structures such as phrases in SMT to achieve a direct connection at the parameter level \cite{wu:07}.

Fortunately, both the source-to-pivot and pivot-to-target models include the word embeddings of the pivot language as parameters. It is possible to connect the two models via pivot word embeddings.

More formally, let $\mathcal{V}_{x \rightarrow z}^{z}$ be the pivot vocabulary of the source-to-pivot model and $\mathcal{V}_{z \rightarrow y}^{z}$ be the pivot vocabulary of the pivot-to-target model. We use $w$ to denote a word in the pivot language and $\bm{\theta}_{x \rightarrow z}^{w} \in \mathbb{R}^{d}$ to denote the vector representation of $w$ in the source-to-pivot model. $\bm{\theta}_{z \rightarrow y}^{w} \in \mathbb{R}^{d}$ is defined in a similar way.

Our first connection term encourages the two models to generate the same vector representations for pivot words in the intersection of two vocabularies:
\begin{eqnarray}
&&\mathcal{R}_{\mathrm{hard}}(\bm{\theta}_{x \rightarrow z}, \bm{\theta}_{z \rightarrow y}) \nonumber \\
&=& \prod_{w \in \mathcal{V}^{z}_{x \rightarrow z} \cap \mathcal{V}^{z}_{z \rightarrow y}} \delta(\bm{\theta}^{w}_{x \rightarrow z}, \bm{\theta}^{w}_{z \rightarrow y})
\end{eqnarray}
where $\delta(\bm{\theta}^{w}_{x \rightarrow z}, \bm{\theta}^{w}_{z \rightarrow y})=1$ if the two vectors $\bm{\theta}^{w}_{x \rightarrow z}$ and $\bm{\theta}^{w}_{z \rightarrow y}$ are identical. Otherwise, $\delta(\bm{\theta}^{w}_{x \rightarrow z}, \bm{\theta}^{w}_{z \rightarrow y})=0$.

As word embeddings seem hardly to be exactly identical due to the divergence of natural languages,  an alternative is to soften the above hard matching constraint by penalizing the Euclidean distance between two vectors:
\begin{eqnarray}
&& \mathcal{R}_{\mathrm{soft}}(\bm{\theta}_{x \rightarrow z}, \bm{\theta}_{z \rightarrow y}) \nonumber \\
&=& - \sum_{w \in \mathcal{V}^{z}_{x \rightarrow z} \cap \mathcal{V}^{z}_{z \rightarrow y}} || \bm{\theta}^{w}_{x \rightarrow z} - \bm{\theta}^{w}_{z \rightarrow y} ||_2
\end{eqnarray}

The third connection term assumes that there is a small bridging source-target parallel corpus $D_{x,y}=\{\langle \mathbf{x}^{(s)}, \mathbf{y}^{(s)} \rangle\}_{s=1}^{S}$ ({\em Bridging Corpus}) available. The connection term is defined as the log-likelihood of the bridging data:
\begin{eqnarray}
&&\mathcal{R}_{\mathrm{likelihood}}(\bm{\theta}_{x \rightarrow z}, \bm{\theta}_{z \rightarrow y}) \nonumber \\
&=& \sum_{s=1}^{S} \log P(\mathbf{y}^{(s)}|\mathbf{x}^{(s)}; \bm{\theta}_{x \rightarrow z}, \bm{\theta}_{z \rightarrow y}) \\
&=& \sum_{s=1}^{S}\log \sum_{\mathbf{z}} P(\mathbf{z}|\mathbf{x}^{(s)}; \bm{\theta}_{x \rightarrow z})P(\mathbf{y}^{(s)}|\mathbf{z}; \bm{\theta}_{z \rightarrow y}) \label{eq:bridge}
\end{eqnarray}

\subsection{Training}

In training, our goal is to find the optimal source-to-pivot and pivot-to-target model parameters that maximize the training objective:
\begin{eqnarray}
\hat{\bm{\theta}}_{x \rightarrow z}, \hat{\bm{\theta}}_{z \rightarrow y} = \argmax_{\bm{\theta}_{x \rightarrow z}, \bm{\theta}_{z \rightarrow y}}\Big\{ \mathcal{J}(\bm{\theta}_{x \rightarrow z}, \bm{\theta}_{z \rightarrow y}) \Big\}
\end{eqnarray}

The partial derivative of $\mathcal{J}(\bm{\theta}_{x \rightarrow z}, \bm{\theta}_{z \rightarrow y})$ with respect to the parameters $\bm{\theta}_{x \rightarrow z}$ of the source-to-pivot model can be calculated as:
\begin{eqnarray}
&& \frac{\partial \mathcal{J}(\bm{\theta}_{x \rightarrow z}, \bm{\theta}_{z \rightarrow y})}{\partial \bm{\theta}_{x \rightarrow z}}   \nonumber \\
&=& \sum_{m=1}^{M} \frac{\partial \log P(\mathbf{z}^{(m)}|\mathbf{x}^{(m)}; \bm{\theta}_{x \rightarrow z})}{\partial \bm{\theta}_{x \rightarrow z}}
+ \nonumber \\
&& \quad \quad \lambda \frac{\partial \mathcal{R}(\bm{\theta}_{x \rightarrow z},\bm{\theta}_{z \rightarrow y})}{\bm{\partial \theta}_{x \rightarrow z}}
\label{eq:derivative}
\end{eqnarray}
The partial derivative with respect to the parameters $\bm{\theta}_{z \rightarrow y}$ can be calculated similarly.

The gradients of the first and second connection terms $\mathcal{R}_{\mathrm{hard}}(\bm{\theta}_{x \rightarrow z},\bm{\theta}_{z \rightarrow y})$ and $\mathcal{R}_{\mathrm{soft}}(\bm{\theta}_{x \rightarrow z},\bm{\theta}_{z \rightarrow y})$ with respect to model parameters are easy to calculate. However, calculating the gradients of the third connection term $\mathcal{R}_{\mathrm{likelihood}}(\bm{\theta}_{x \rightarrow z},\bm{\theta}_{z \rightarrow y})$ involves enumerating all possible pivot sentences in an exponential search space (see Eq. (\ref{eq:bridge})).

To alleviate this problem, we follow standard practice to use a subset to approximate the full space \cite{shen:16,cheng:16b}. Two methods can be used to generate a subset: sampling $k$ translations from the full space \cite{shen:16} or generating a top-$k$ list of candidate translations \cite{cheng:16b}. We find that using top-$k$ lists leads to better results than sampling in our experiments.


We use standard mini-batched stochastic gradient descent algorithms to optimize model parameters. In each iteration, three mini-batches are constructed by randomly selecting sentence pairs from the source-pivot parallel corpus $D_{x,z}$, the pivot-target parallel corpus $D_{z,y}$, and the bridging source-target parallel corpus $D_{x,y}$ (only available for the third connection term), respectively. After separate gradient calculation in each mini-batch, the gradients are collected to update model parameters.

\begin{table}[!t]
\centering
\begin{tabular}{|l|l|l||r|r|}
\hline
Corpus & Lang.& & Source & Target \\
\hline \hline
\multirow{9}{*}{Europarl} & & \# Sent. & \multicolumn{2}{c|}{850K} \\
 \cline{3-5}
 & es-en  & \# Word & 22.32M & 21.44M \\
 \cline{3-5}
 & & Vocab. & 118.81K & 78.37K \\
 \cline{2-5}
 & & \# Sent. &  \multicolumn{2}{c|}{840K} \\
 \cline{3-5}
 & de-en  & \# Word & 20.88M & 21.91M \\
 \cline{3-5}
 & & Vocab. & 242.87K & 80.44KM \\
  \cline{2-5}
 & & \# Sent. &  \multicolumn{2}{c|}{900K} \\
 \cline{3-5}
 & en-fr  & \# Word & 22.56M & 25.00M \\
 \cline{3-5}
 & & Vocab. & 80.08K & 98.50K \\
 \hline \hline
\multirow{6}{*}{WMT} & & \# Sent. & \multicolumn{2}{c|}{6.78M} \\
 \cline{3-5}
 & es-en & \# Word &183.01M & 166.28M \\
 \cline{3-5}
 & & Vocab. & 0.98M &  0.91M \\
  \cline{2-5}
 & & \# Sent. &  \multicolumn{2}{c|}{9.29M} \\
 \cline{3-5}
 & en-fr & \# Word & 227.06M & 258.95M \\
 \cline{3-5}
 & & Vocab. & 0.23M & 1.19M \\
 \hline
\end{tabular}
\caption{Characteristics of Spanish-English, German-English and English-French datasets on the Europarl and WMT corpora. ``es'' denotes Spanish, ``en'' denotes English, ``de'' denotes German, and ``fr'' denotes French.}
\label{table:data}
\end{table}

\begin{table*}[!t]

\centering
\begin{tabular}{|c|c|l|l|l|l|l|l|l|}
\hline
\multirow{2}{*}{Training} & \multirow{2}{*}{Connection} & \multirow{2}{*}{Dataset} & \multicolumn{3}{|c|}{Spanish-French } &\multicolumn{3}{|c|}{German-French }   \\
\cline{4-9}
& & &es$\rightarrow$ en &en $\rightarrow$ fr & es $\rightarrow$ fr  &de$\rightarrow$ en &en $\rightarrow$ fr & de $\rightarrow$ fr   \\
\hline \hline
\multirow{2}{*}{indep.}       & \multirow{2}{*}{-} & Dev. & 31.53 &30.46 &29.52 &26.52 &30.46 &23.67 \\
 &  & Test &31.54 &31.42 &29.79 &26.47 &31.42 &23.70 \\
\hline \hline
\multirow{6}{*}{joint} & \multirow{2}{*}{hard} & Dev. &31.81 &30.18 &29.11 &26.48 &30.47 &23.87 \\
&  & Test &31.55 &31.13 &29.93 &26.58 &31.35 &23.88 \\
\cline{2-9}
& \multirow{2}{*}{soft} & Dev. &32.11$^{**}$ &30.41 &30.24$^{**}$ &26.92 &30.39 &23.99 \\
  &  & Test &31.96$^{*}$ &31.40 &30.57$^{**}$ &26.55 &31.33 &23.79 \\
\cline{2-9}
& \multirow{2}{*}{likelihood} & Dev. &33.35$^{**}$ &31.63$^{**}$ &32.45$^{**}$ &27.90$^{**}$ &31.49$^{**}$ &25.21$^{**}$ \\
&  & Test &33.54$^{**}$ &32.33$^{**}$ &32.59$^{**}$ &28.01$^{**}$ &32.34$^{**}$ &25.93$^{**}$ \\
\hline
\end{tabular}
\caption{Comparison between independent and joint training on Spanish-French and German-French translation tasks using the Europarl corpus. English is treated as the pivot language.
The BLEU scores are case-insensitive. ``*'': significantly better than independent training ($p<0.05$); ``**'': significantly better than independent training  ($p<0.01$). We use the statistical significance test with paired bootstrap resampling \protect \cite{koehn:04}.}
\label{table:comparison_eu}
\end{table*}

\section{Experiments}

\subsection{Setup}

We evaluated our approach on two translation tasks:
\begin{enumerate}
\item Spanish-English-French: Spanish as the source language, English as the pivot language, and French as the target language,
\item German-English-French: German as the source language, English as the pivot language, and French as the target language.
\end{enumerate}

Table \ref{table:data} shows the statistics of the Europarl and WMT corpora used in our experiments. We use \verb|tokenize.perl| script for tokenization. For each language pair, we remove the empty lines and retain sentence pairs with no more than 50 words. To avoid the intersection of the source-pivot and pivot-target corpora, we split the overlapped pivot-language sentences of source-to-pivot and pivot-to-target corpora into two separate parts with equal size and
merge them separately with the non-overlapping parts for each language pair.

\begin{figure}[!t]
\centering
\includegraphics[width=0.50\textwidth]{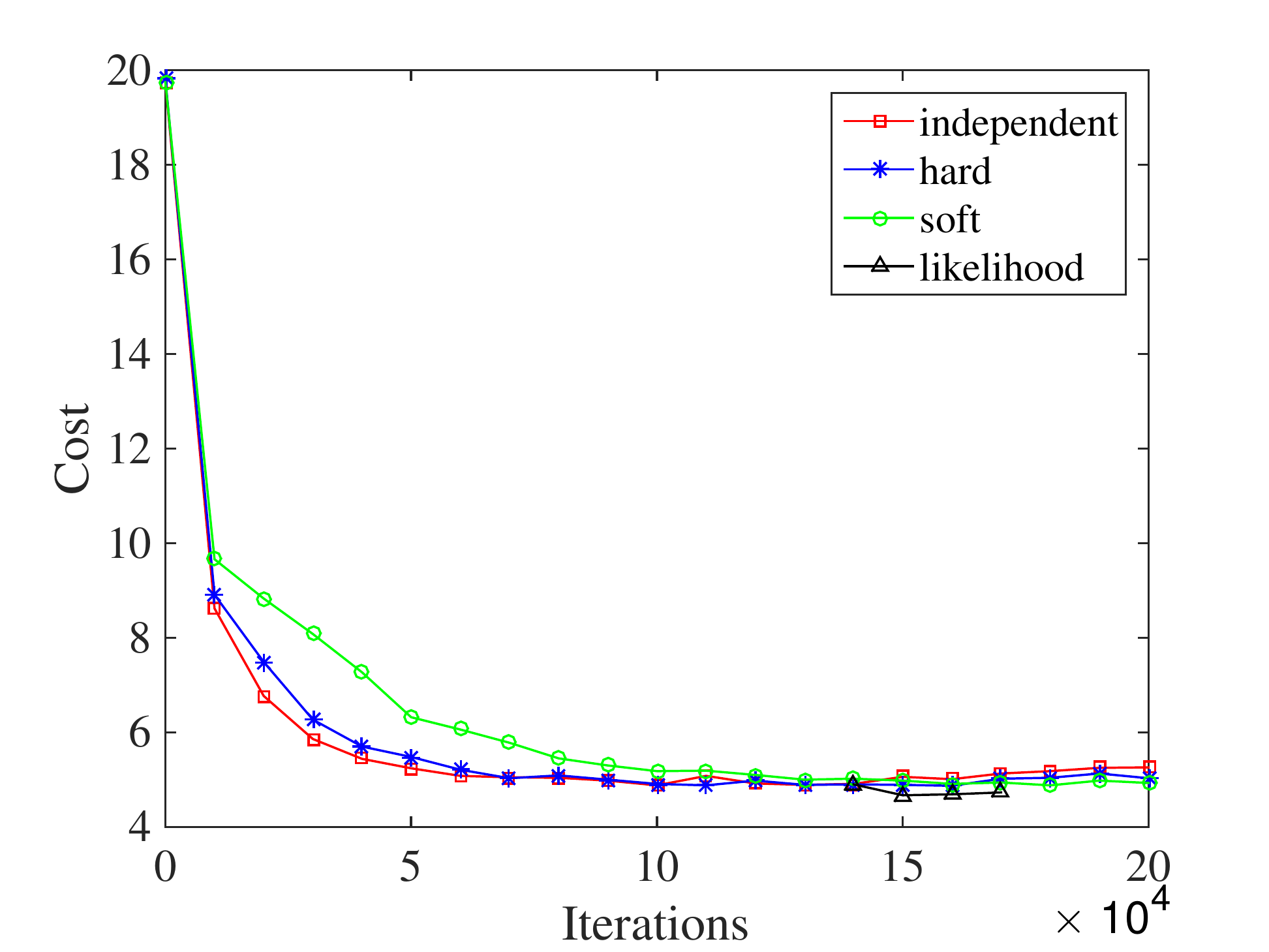}
\caption{Learning curves of independent training and joint training on different connection terms.} \label{fig:learn}
\end{figure}

\begin{table*}[!t]
\centering
\small
\begin{tabular}{l|l|p{1.5\columnwidth}}
\hline
\multirow{3}{*}{\textproc{GroundTruth}}  &source & uno no debe empezar a dudar en p\'ublico del valor , tampoco del valor inmediato en el aspecto material , de esta ampliaci\'on . \\
\cline{2-3} &pivot & it makes little sense to start to doubt in public the value , including the direct value at a material level , of this enlargement .\\
\cline{2-3} &target & il ne faut pas commencer \`a douter en public de la valeur , ni de la valeur imm\'ediate , de la port\'ee mat\'erielle de cet \'elargissement . \\
\hline
\multirow{2}{*}{\textproc{Indep.}}  &pivot & one should not begin \textcolor{blue}{\textit{to doubt in}} terms of \textcolor{blue}{\textit{the value}} of courage , or of the immediate effect on material ,  of \textcolor{blue}{\textit{enlargement .}}  [BLEU: 13.33] \\
\cline{2-3} &target & \textcolor{blue}{\textit{il ne}} faudrait pas se tromper en termes de valeur de courage ou d ' effet imm\'ediat sur le mat\'eriel , l ' \textcolor{blue}{\textit{\'elargissement .}}  [BLEU: 8.69]\\
\hline
\multirow{2}{*}{\textproc{Hard}}  &pivot & one must not \textcolor{blue}{\textit{start to doubt in}} the public , not the immediate value in the material , \textcolor{blue}{\textit{this enlargement .}}  [BLEU: 19.02] \\
\cline{2-3} &target &\textcolor{blue}{\textit{il ne faut pas}} que l ' on \textcolor{blue}{\textit{commence \`a douter}} , ni au public , ni \`a la \textcolor{blue}{\textit{valeur imm\'ediate ,}} \`a l ' \textcolor{blue}{\textit{\'elargissement .}} [BLEU: 25.36] \\
\hline
\multirow{3}{*}{\textproc{Soft}}  &pivot & one cannot start thinking of the value of \textcolor{blue}{\textit{the value ,}} and the immediate courage \textcolor{blue}{\textit{, of this enlargement .}} [BLEU: 21.57] \\
\cline{2-3} &target & on \textcolor{blue}{\textit{ne peut pas commencer \`a}} penser \`a la valeur \textcolor{blue}{\textit{de la valeur}} , au courage imm\'ediat , \textcolor{blue}{\textit{de cet \'elargissement .}} [BLEU: 26.60] \\
\hline
\multirow{3}{*}{\textproc{Liklihhod}}  &pivot & one must not \textcolor{blue}{\textit{start to}} question the value of \textcolor{blue}{\textit{the value ,}} either of the immediate value in the material aspect \textcolor{blue}{\textit{, of this enlargement .}} [BLEU: 24.60]\\
\cline{2-3} &target & \textcolor{blue}{il ne faut pas commencer \`a} remettre en question la valeur \textcolor{blue}{\textit{de la valeur , ni de la valeur imm\'ediate}} de l ' aspect mat\'eriel , \textcolor{blue}{\textit{ de cet \'elargissement .}} [BLEU: 56.40]\\
\hline
\end{tabular}

\caption{Examples of  pivot and target translations using the pivot-based translation strategy.
We observe that our approaches generate better translations for both pivot and target sentences.
We italicize \protect \textcolor{blue}{\textit{correct translation segments}} which are no short than 2-grams.} \label{table:example}

\label{table:example}
\end{table*}

The Europarl corpus consists of 850K Spanish-English sentence pairs with 22.32M Spanish words and 21.44M English words, 840K German-English sentence pairs with 20.88M German words and 21.91M English words, and 900K English-French sentence pairs with 22.56M English words and 25.00M French words.  The WMT 2006 shared task datasets are used as the development and test sets. The evaluation metric is case-insensitive BLEU \cite{Papineni:02} as calculated by the \verb|multi-bleu.perl| script.

The WMT corpus is composed of the Common Crawl, News Commentary, Europarl v7 and UN corpora. The Spanish-English parallel corpus consists of 6.78M sentence pairs with 183.01M Spanish words and 166.28M English words. The English-French parallel corpus comprises 9.29M sentence pairs with 227.06M English words and 258.95M French words. The {\em newstest2011} and {\em newstest2012} datasets serve as development and test sets. We use case-sensitive BLEU as the evaluation metric.

We use the attention-based neural machine translation system \textproc{RNNsearch} \cite{Bahdanau:15} in our experiments. For the Europarl corpus in Table \ref{table:data}, we set the vocabulary size of all the languages to 30K which covers over 99\% of words for English, Spanish and French and over 97 \% for German. We follow Jean et al. \shortcite{Jean:15} to address rare words. For Spanish-English and English-French corpora from the WMT corpus, due to large vocabulary size, we adopt byte pair encoding \cite{sennrich:16b} to split rare words into sub-words. The size of sub-words is set to 43K, 33K, 43K respectively for Spanish, English, and French. These sub-words cover 100\% of the text.

We set the hyper-parameter $\lambda$ for balancing between likelihood and the connection term to 1.0. The threshold of gradients is set to 0.1. The bridging source-target parallel corpus contains 100K sentence pairs that do not overlap with the training data. We set $k$ to 10 for calculating top-$k$ lists to approximate the full search space. The parameters for the source-to-pivot and pivot-to-target translation models in the likelihood connection term are initialized by pre-trained model parameters.



\begin{table}[!t]

\centering
\begin{tabular}{|c|c|l|l|l|}
\hline
\multirow{2}{*}{Method} &\multirow{2}{*}{Dataset}& \multicolumn{3}{|c|}{Spanish-French (WMT)}   \\
\cline{3-5}
& &es$\rightarrow$ en &en $\rightarrow$ fr & es $\rightarrow$ fr   \\
\hline \hline
\multirow{2}{*}{indep.}&Dev. &27.62 &27.90 &24.92  \\
&Test &29.03 &25.82 &24.60\\
\hline
\multirow{2}{*}{likelihood} &Dev. &28.92$^{**}$ &28.52$^{**}$ &26.24$^{**}$  \\
 &Test &30.43$^{**}$ &26.36$^{**}$ & 25.78$^{**}$\\
\hline
\end{tabular}
\caption{Results on Spanish-French translation task from WMT corpus. English is treated as the pivot language.  ``**'': significantly better than independent training  ($p<0.01$).}
\label{table:comparison_wmt}
\end{table}

\begin{table}[!t]
\centering
\begin{tabular}{|c|l|l|}
\hline
Systems &{\em newstest2012} &{\em newstest2013}  \\
\hline \hline
 Firat et al. \shortcite{firat:16}  &21.81 &21.46 \\
\hline
{\em this work} &25.95$^{**}$  &25.78$^{**}$ \\
\hline
\end{tabular}
\caption{Comparison with Firat et al. \protect \shortcite{firat:16}.  ``**'': significantly better than independent training  ($p<0.01$).}
\label{table:comparison_firat}
\end{table}

\subsection{Results on the Europarl Corpus}

Table \ref{table:comparison_eu} shows the comparison results between our joint training on three connection terms and independent training on the Europarl Corpus. For the source-to-target translation task, we
present source-to-pivot, pivot-to-target and source-to-target translation results compared with independent training. In Spanish-to-French translation task,
soft connection achieves significant improvements in Spanish-to-French and Spanish-to-English directions although hard connection still performs comparably with independent training.
In German-to-French translation task, soft and hard connections also achieve comparable performances with independent training.

In contrast, we find that likelihood connection dramatically improves
translation performance on both Spanish-to-French and German-to-French corpora (up to +2.80 BLEU scores in Spanish-to-French and up to 2.23 BLEU scores in German-to-French). The significant improvements for source-to-pivot and pivot-to-target directions
are also observed. This suggests that introducing source-to-target parallel corpus to maximize $P(\mathbf{y}|\mathbf{x};\bm{\theta}_{x \rightarrow z},\bm{\theta}_{z \rightarrow y})$ with $\mathbf{z}$
as latent variables makes the source-to-pivot and pivot-to-target
translation models improved collaboratively. 

Table \ref{table:example} shows pivot and target translation examples of independent training and our approaches. Apparently, our approaches improve
translation quality of both pivot sentences and target sentences.

According to Eq. (\ref{eq:source-to-target}), the cost of the source-to-target
model can be decomposed into the cost of source-to-pivot and pivot-to-target models. Because we have a small test trilingual corpus, (Spanish, English, French), we use the English sentence
to approximate the latent variables in Eq. (\ref{eq:source-to-target}). Then we calculate the cost of Spanish-to-French on the trilingual corpus. Figure \ref{fig:learn} shows the learning
 curves of the test cost of independent training and joint training on three connection terms. 
We can find that hard and soft connections learn slower than the independent training. 
Likelihood connection drives its cost lower after fine-tuning based on pre-trained parameters in just 10K iterations.

\begin{table}[!t]

\centering
\begin{tabular}{|l|l|c|c|}
\hline
Corpus  &Lang. &source-target & source-pivot-target    \\
\hline \hline
\multirow{2}{*}{Europarl}&es $\rightarrow$ fr  &26.37 &29.79 \\
\cline{2-4}
&de $\rightarrow$ fr &14.02 & 23.70\\
\hline
WMT &es $\rightarrow$ fr  &11.75 &24.60 \\
\hline
\end{tabular}
\caption{Translation performance on bridging corpora.}
\label{table:little_nmt}
\end{table}

\begin{table}[!t]

\centering
\begin{tabular}{|r|c|c|c|}
\hline
\# Sent.  &es$\rightarrow$ en &en $\rightarrow$ fr & es $\rightarrow$ fr   \\
\hline \hline
0 &31.53  &30.46  &29.52 \\
\hline
1K &32.64 &30.29 &30.23  \\
\hline
10K &32.92 &30.93 &31.51  \\
\hline
50K &33.29 &31.57 &32.40 \\
\hline
100K &33.35 &31.63 &32.45 \\
\hline
\end{tabular}
\caption{Effect of the data  size of source-to-target parallel corpora ({\em Bridge Corpora}) used in \textproc{Likelihood}.} 
\label{table:diff_size}
\end{table}

\subsection{Results on the WMT Corpus}

Likelihood connection obtains the best performance in our three proposed connection terms according to experiments on the Europarl corpus.
To further verify its practicability, Table \ref{table:comparison_wmt} shows results on the WMT corpus
which is a much larger corpus. We find that likelihood connection  still outperforms independent training significantly on Spanish-to-English, English-to-French and Spanish-to-French directions (up to +1.18 BLEU scores in Spanish-to-French).

We also compare our approach with Firat et al. \shortcite{firat:16}. They propose a multi-way, multilingual NMT model to build a source-to-target translation model. Although our parallel training corpus is much smaller
 than theirs, Table \ref{table:comparison_firat} shows that our approach achieves substantial improvements over them (up to +4.32 BLEU).

\subsection{Effect of Bridging Corpora}

As bridging corpora are used in likelihood connection term for ``bridging" the source-to-pivot and pivot-to-target translation models, why do not we directly build NMT systems
with these corpora?

We train source-to-target models using bridging corpora and show translation results
in Table \ref{table:little_nmt} . We observe that performance is much worse than that in Table \ref{table:comparison_eu} and Table \ref{table:comparison_wmt} using the pivot-based
translation strategy. It indicates that NMT yields poor performance on low-resource languages and the pivot-based translation strategy remedies the drawback to alleviate data scarcity effectively.

We also investigate the effect of the data size of bridging corpora on the likelihood connection. Table \ref{table:diff_size} shows that using a small parallel corpus (1K sentence pairs)
has made a measurable improvement. When more than 50K sentence pairs are added, the further improvements become modest.
This finding suggests that a small corpus suffices to enable the likelihood connection to reach the reasonable performance.

\section{Related Work}

Our work is inspired by two lines of research: (1) machine translation with pivot languages and (2) incorporating additional data resource for NMT.

\subsection{Machine Translation with Pivot Languages}
Machine translation suffers from the scarcity of parallel corpora. For low-resource language pairs, a pivot language is introduced to ``bridge" source and target languages in statical
machine translation \cite{cohn:07,wu:07,utiyama:07,zahabi:13,el:13}. 

In NMT, Firat et al. \shortcite{firat:16} and Johnson et al \shortcite{johnson:16}  propose multi-way, multilingual NMT models that enable zero-resource machine translation.
They also need to apply pivot-based approaches into NMT to ameliorate the performance of zero-resource machine translation.
Zoph et al. \shortcite{zoph:16} adopt transfer learning to fine-tune parameters of the low-resource language pairs using trained parameters on the high-resource language pairs.
However, our approach aims to jointly train source-to-pivot and pivot-to-target NMT models, which can alleviate the error propagation of pivot-based approaches. We use connection terms to ``bridge" these two models and make them benefit each other.

\subsection{Incorporating Additional Data Resources for NMT}

Due to the limit in quantity, quality and coverage for parallel corpora,  additional data resources have raised attention recently. Gulccehre et al \shortcite{Gulcehre:15} propose to incorporate target-side monolingual corpora as a language model for NMT.  Sennrich, Haddow, and Birch \shortcite{sennrich:16a} pair the target monolingual corpora with its corresponding
translations, then merge them with parallel corpora for retraining source-to-target model.
Zhang and Zong \shortcite{zhang:16} propose two approaches,  self-training algorithm and multi-task learning framework, to incorporate source-side monolingual corpora.
Cheng et al. \shortcite{cheng:16b} introduce an autoencoder framework to reconstruct monolingual sentences using source-to-target and target-to-source NMT models. The proposed
model can exploit both source and target monolingual corpora. In contrast to Cheng et al. \shortcite{cheng:16b}, the objective of our likelihood connection is to maximize the probability of target-language sentences through pivot sentences given source sentences. We use a small source-to-target parallel corpus to train
source-to-pivot and pivot-to-target NMT models jointly.
\section{Conclusion}

We present joint training for pivot-based neural machine translation.
Experiments on different language pairs confirm that our approach achieves significant improvements. It is appealing to combine source
and pivot sentences for decoding target sentences \cite{firat:16} or train a multi-source model directly \cite{zoph:16b}. We also plan to study better connection terms for our joint training.


\bibliographystyle{named}
\bibliography{ijcai17}

\end{document}